\def\year{2021}\relax
\documentclass[letterpaper]{article} 
\usepackage{aaai21}  
\usepackage{times}  
\usepackage{helvet} 
\usepackage{courier}  
\usepackage[hyphens]{url}  
\usepackage{graphicx} 
\usepackage{natbib}  
\usepackage{caption} 
\usepackage{caption}
\usepackage{subcaption}
\usepackage{microtype}
\usepackage{graphicx}
\usepackage{color}
\usepackage{tabularx}
\usepackage[switch]{lineno}

\title{Explicitly Modeled Attention Maps for Image Classification}

\title{Explicitly Modeled Attention Maps for Image Classification}
\author {
        Andong Tan\thanks{Equal contribution} \textsuperscript{\rm 1,\rm4},
        Duc Tam Nguyen\footnotemark[1] \textsuperscript{\rm 2,\rm4},
        Maximilian Dax\textsuperscript{\rm 3,\rm4},
        Matthias Nie{\ss}ner\textsuperscript{\rm 1},
        Thomas Brox\textsuperscript{\rm 2} \\
}
\affiliations {
    \textsuperscript{\rm 1} Technical University of Munich \\
    \textsuperscript{\rm 2} University of Freiburg \\
    \textsuperscript{\rm 3} University of Bonn \\
    \textsuperscript{\rm 4} Robert Bosch GmbH\\
    andong.tan@tum.de, nguyen@cs.uni-freiburg.de, maximiliandax@gmx.com, niessner@tum.de, brox@cs.uni-freiburg.de
}
\begin{document}
\maketitle

\newcommand{\Softmax}{\text{Softmax}}
\newcommand{\Concat}{\text{Concat}}
\newcommand{\Multihead}{\text{MultiHead}}
\newcommand{\PosEncoding}{\text{Pos.Encoding}}
\newcommand{\Att}{Att}
\newcommand{\Conv}{Conv}
\newcommand{\ExpAtt}{ExpAtt}
\newcommand{\Norm}{Norm}

\newcommand\MATTHIAS[1]{\textcolor{red}{#1}}
\newcommand\tam[1]{\textcolor{blue}{#1}}
\newcommand\andong[1]{\textcolor{cyan}{#1}}
\newcommand\maxd[1]{\textcolor{magenta}{#1}}
\newcommand{\theHalgorithm}{\arabic{algorithm}}

\begin{abstract}
Self-attention networks have shown remarkable progress in computer vision tasks such as image classification. 
The main benefit of the self-attention mechanism is the ability to capture long-range feature interactions in attention-maps. However, the computation of attention-maps requires a learnable key, query, and positional encoding, whose usage is often not intuitive and computationally expensive. 
To mitigate this problem, we propose a novel self-attention module with explicitly modeled attention-maps using only a single learnable parameter for low computational overhead. 
The design of explicitly modeled attention-maps using geometric prior is based on the observation that the spatial context for a given pixel within an image is mostly dominated by its neighbors, while more distant pixels have a minor contribution. 
Concretely, the attention-maps are parametrized via simple functions (e.g., Gaussian kernel) with a learnable radius, which is modeled independently of the input content. 
Our evaluation shows that our method achieves an accuracy improvement of up to 2.2\%  over the ResNet-baselines in ImageNet ILSVRC and outperforms other self-attention methods such as AA-ResNet152 in accuracy by 0.9\% with 6.4\% fewer parameters and 6.7\% fewer GFLOPs. This result empirically indicates the value of incorporating geometric prior into self-attention mechanism when applied in image classification.
\end{abstract}

\section{Introduction}

\begin{figure*}
\centering
\begin{subfigure}[b]{.17\textwidth}
    \centering
    \includegraphics[width=.99\linewidth]{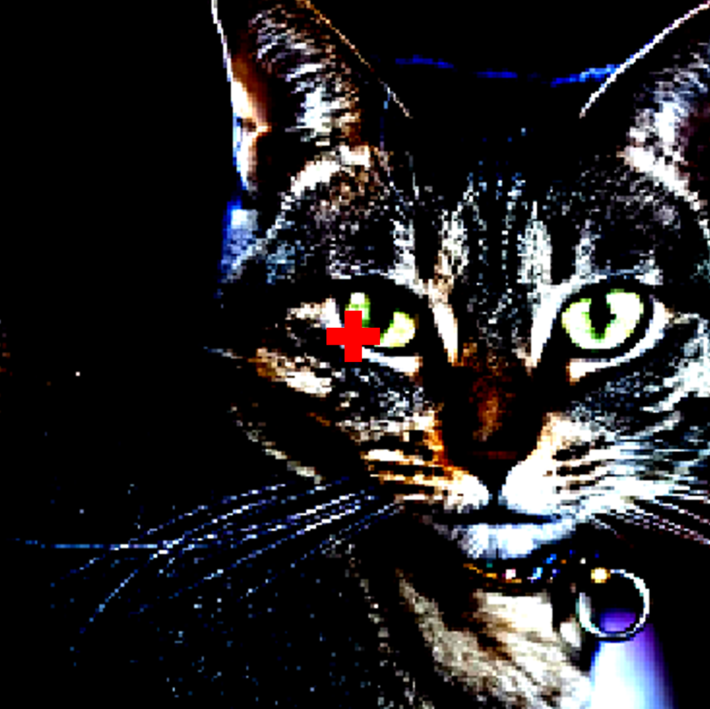}
    \caption{}
    \label{fig:cat_origin_redcross_center}
\end{subfigure}
\begin{subfigure}[b]{.17\textwidth}
    \centering
    \includegraphics[width=.99\linewidth]{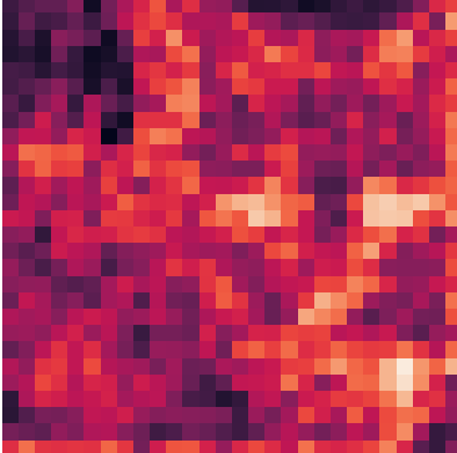}
    \caption{}
    \label{fig:normkq405}
\end{subfigure}
\begin{subfigure}[b]{.17\textwidth}
    \centering
    \includegraphics[width=.99\linewidth]{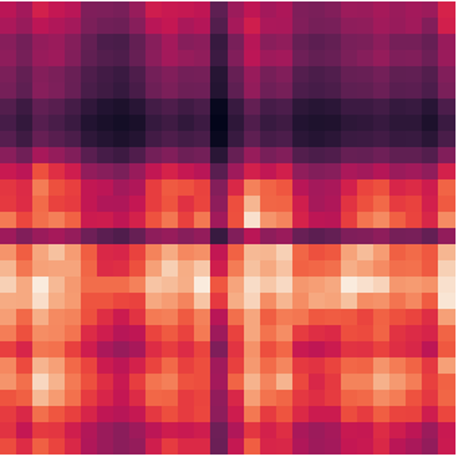}
    \caption{}
    \label{fig:normkqp405}
\end{subfigure}
\begin{subfigure}[b]{.17\textwidth}
    \centering
    \includegraphics[width=.99\linewidth]{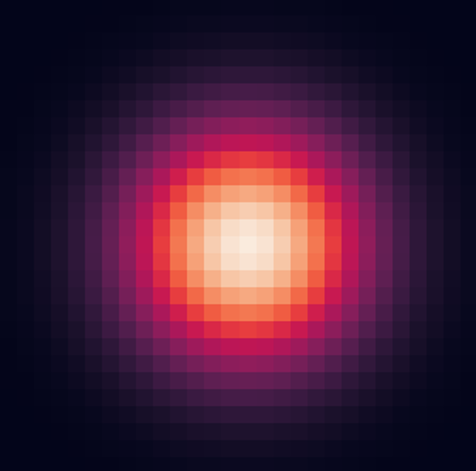}
    \caption{}
    \label{fig:gaussiannorm405}
\end{subfigure}\\
\begin{subfigure}[b]{.17\textwidth}
    \centering
    \includegraphics[width=.99\linewidth]{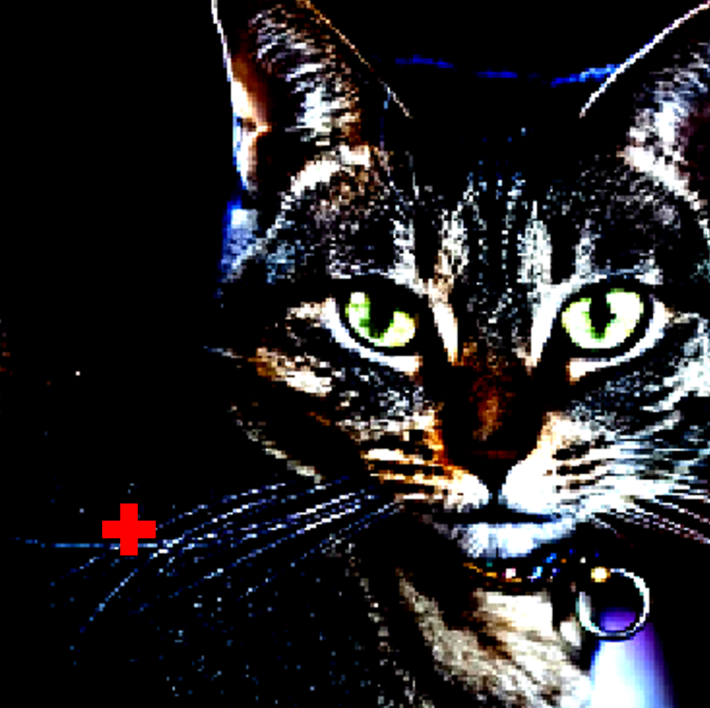}
    \caption{}
    \label{fig:cat_origin_redcross_left}
\end{subfigure}
\begin{subfigure}[b]{.17\textwidth}
    \centering
    \includegraphics[width=.99\linewidth]{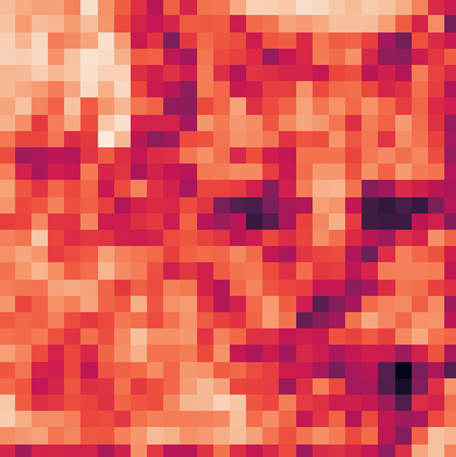}
    \caption{}
    \label{fig:normkq592}
\end{subfigure}
\begin{subfigure}[b]{.17\textwidth}
    \centering
    \includegraphics[width=.99\linewidth]{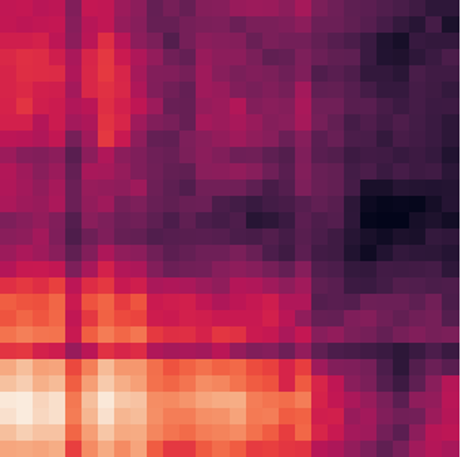}
    \caption{}
    \label{fig:normkqp592}
\end{subfigure}
\begin{subfigure}[b]{.17\textwidth}
    \centering
    \includegraphics[width=.99\linewidth]{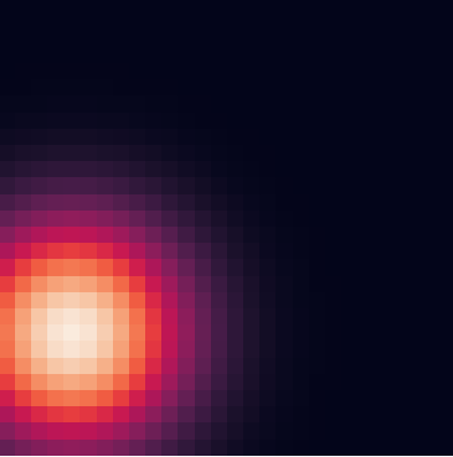}
    \caption{}
    \label{fig:gaussiannorm592}
\end{subfigure}
\caption{Attention-maps of two pixels: (a,e) are the input from ImageNet, red crosses indicate the positions of 2 pixels. (b,f) show key$\times$query; (c,g) show key$\times$query+positional encoding; (d,h) show explicitly modeled attention-maps using Gaussian kernel. All pictures are extracted from the first layer where attention mechanism is applied in ResNet50. Brighter color indicates higher weight. First and second row of pictures correspond to attention-maps of the center and left pixels. All values are normalized for visualization.
}
\label{fig:attention_maps_example}
\end{figure*}

An attention mechanism allows the network to focus on the global context in each layer. Attention is an essential part of human vision, also known as foveation, which allows the vision system to focus its limited resources on a small part of the input signal. The implementation of this concept as self-attention in the transformer network \citep{vaswani2017attention} has resulted in a substantial performance increase in natural language processing \citep{devlin2018bert, radford2018improving}. Recent works in computer vision \citep{bello2019attention, parmar2019stand} proposed self-attention for object recognition tasks.
Their suggested modification of common network architectures, such as the Resnet \citep{he2016deep,he2016identity}, leads to significant performance improvement over the original convolutional baseline. 

One key benefit of a self-attention layer is that such a layer can incorporate the entire spatial context for the computation of features in the subsequent layers through the calculation of attention-maps. As an intuitive explanation, attention-maps express how much attention different areas of the input receive when focusing on a particular pixel position. However, the high degree of freedom of self-attention networks, such as in \citep{bello2019attention}, requires learning the weights for the key and the query to calculate attention-maps. There is no obvious need for this excessive, computationally expensive parametrization in the context of computer vision tasks. By visualizing the weights, we show that the content-dependent key and query play a minor role in the final attention-maps, where an area with high attention weight is more related to the geometric position; compare Fig. \ref{fig:normkqp405}, \ref{fig:normkqp592} to Fig. \ref{fig:normkq405}, \ref{fig:normkq592}. Further, in vision tasks such as image classification, a common observation is that neighbor pixels are more related than distant ones. In other words, when focusing on particular pixels, neighbor areas should receive higher attention.

Based on these insights, we propose a self-attention mechanism with explicitly modeled attention-maps, which considers the global context information with positive correlation to the distance between pixels. The freedom of the attention module is reduced to a predefined form (e.g., Gaussian kernel) with a learnable parameter, as illustrated in Fig.~\ref{fig:attmap}. We integrate the local context prior in self-attention explicitly by restricting the learnable attention-map to a centered, yet global kernel with a learnable shape parameter. The primary motivation is to maintain global information for the feature computation while reducing the freedom caused by learning key and query for higher efficiency. Surprisingly, networks augmented with such modules not only outperform regular self-attention in parameters, memory and computation cost, but also achieve very competitive accuracy.

The contributions of this paper are
summarized as follows:
(1) We analyze the efficiency of AA-Net \citep{bello2019attention} and empirically show that geometric information plays an essential role in attention-maps.
(2) Based on the above analysis, we propose a novel self-attention module with explicitly modeled attention-maps under the assumption that neighbor pixels are more related than distant ones. We investigate fixed attention-maps parametrized by different global, yet centered simple functions (e,g, cosine, linear) to model monotonically decreased attention paid to distant pixels w.r.t. centered pixel. Further, we study the effect of automatically determining attention-maps using the Gaussian kernel with a learnable radius.
(3) Experimental results in CIFAR10, CIFAR100, Tiny ImageNet, and ImageNet show that convolutional networks augmented with such modules have lower model complexity than augmented with regular self-attention while achieving very competitive accuracy.

\section{Related Works}
The most important works on attention in deep networks comprise advances for sequence-to-sequence modeling in natural language processing (NLP) tasks such as neural machine translation \citep{bahdanau2014neural}. More recently, multi-head self-attention \citep{vaswani2017attention, so2019evolved} allows effective pretraining for many NLP-tasks using language modeling as a self-supervision task (\citep{devlin2018bert,radford2018improving}) and for other tasks \citep{Shaw2018SelfAttentionWR, zhang2018self, yu2018qanet, zhang2016gaussian}. Since self-attention is computationally expensive, there are also works exploring efficient self-attention \citep{Shen2018BiDirectionalBS, Shen2019TensorizedSE,kitaev2019reformer}. Further, Synthesizer \citep{Tay2020SynthesizerRS} challenges the necessity of computationally expensive key-query self-attention, and fixed self-attention patterns are proposed for machine translation \citep{Raganato2020FixedES}. However, these proposed self-attention mechanisms are designed for sequences-to-sequence tasks and do not necessarily transfer to imaging tasks with different dynamics. Contrary, the focus of our work is to study the self-attention concept from NLP for learning on computer vision tasks.

Attention methods in image recognition tasks can be roughly categorized into channel attention, spatial attention, or a combination of them. A representative work exploring channel attention is SE-Net \citep{hu2018squeeze}, which calculates channel attention by using global average pooling and channel scaling. GE-Net \citep{hu2018gather} uses depth-wise convolution to calculate spatial attention. CBAM \citep{woo2018cbam} extends SE-Net by additionally considering spatial attention independently. ResNeSt \citep{Zhang2020ResNeStSN} uses a cardinal group to generalize prior work in channel attention. Further, GSoP \citep{gao2019global} exploits channel and spatial attention respectively from a statistical perspective. More recent work such as AA-Net \citep{bello2019attention} calculates spatial and channel attention jointly using the self-attention concept from NLP. These works mainly aim at improving performance with intricate module design. In the domain of efficient attention mechanism, ECA-Net \citep{Wang2019ECANetEC} improves SE-Net \citep{hu2018squeeze} for an efficient channel attention mechanism by controlling the size of 1D convolution.
Different from all the above, we try to improve AA-Net \citep{bello2019attention} and aim at offering efficient attention in spatial and channel dimensions jointly by incorporating geometric prior. 

Overall, to the best of our knowledge, we are the first to propose a self-attention module with explicitly modeled attention-maps in an extremely simplified way for vision task. In this form, the attention-maps are shared across multiple heads and are parameterized by only one single learnable parameter in each layer. Compared to AA-Net \citep{bello2019attention}, our module does not employ any key or query and retains only the value.
Since our method is strongly motivated by AA-Net \citep{bello2019attention}, we first introduce how AA-Net \citep{bello2019attention} applies the self-attention concept from NLP in vision task.

\section{Background on Multi-head Self-attention in Computer Vision \label{attention}}

We first denote the notations used in this section. Following the convention, we denote H, W, C as height, width, and the number of channels of input $X'\in R^{H\times W\times C}$ (Batch dimension is omitted for simplicity). The flattened input is denoted as $X\in R^{HW\times C}$. Further, we define $d$ as the depth of key or query, $d_v^h$ as the depth of value in each head, and $N$ as the total number of heads. 

\begin{figure*}[t]
    \centering
    \includegraphics[width=.8\linewidth]{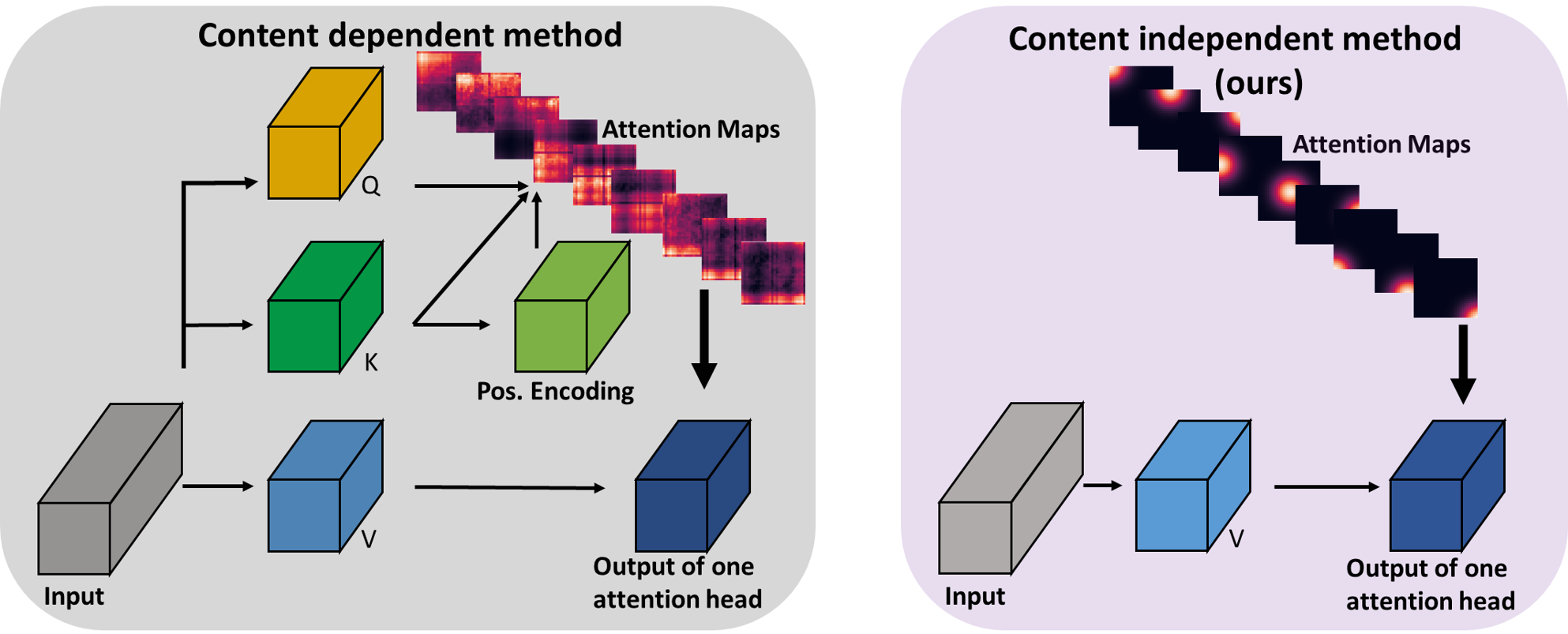}
    \caption{Comparison between self-attention mechanism using content-dependent and independent attention-maps: content-dependent attention-maps are constructed using linearly projected input key (K) and query (Q), while our explicitly modeled attention-maps simply incorporate geometric prior. In the end, the attention-maps are used to calculate the weighted average of the value (V) for the output of one attention head. Attention-maps are restructured for better visualization. 
    }.
    \label{fig:attmap}

\end{figure*}

The multi-head self-attention is calculated as in the  Transformer architecture \citep{vaswani2017attention}. The three steps of the process are defined as \citep{bello2019attention}: %

\begin{equation}
    \label{eq:Att_wo}
    Att(X)=Softmax\left(\frac{K Q^T+PosEncoding}{\sqrt{d}}\right)V
\end{equation}
\begin{equation}
    \label{eq:Att_v}
    V=XW_v^h; K=XW_k^h; Q=XW_q^h
\end{equation}
\begin{equation}
    \label{eqn:multihead}
    Multihead(X)=Concat[Att_1,...,Att_N]W^o
\end{equation}
where $W_v^h \in R^{C \times d_v^h}$, $W_k^h \in R^{C \times d}$, $W_q^h \in R^{C \times d}$, and $W^o \in R^{N d_v^h \times N d_v^h}$ are 4 learnable matrices to calculate value $V$, key $K$, query $Q$ and final output respectively. The positional encoding term refers to a learnable relative positional encoding \citep{shaw-etal-2018-self}, which is translational invariant. The division of $\sqrt{d}$ is designed for better training. The calculations in  Eq. \ref{eq:Att_v} %
are repeated multiple times with different learnable matrices to get multiple $Att(X)$ (also named as one head). In the last step, the results of N heads are concatenated along the depth dimension and linearly projected to achieve final multi-head self-attention. An overview of this method is offered in the left part of Fig. \ref{fig:attmap}%

\paragraph{What are attention-maps?}
The attention-map describes how much attention every pixel in the input is paid to when the model is focusing on one specific pixel. Every pixel has one attention-map, so there are in total HW attention-maps for the input of height $H$ and width $W$, and every attention-map has spatial shape $H\times W$. Therefore the softmax part in Eq. \ref{eq:Att_wo} with the shape $HW\times HW$ indicates the attention paid to all $H \times W$ pixels when focusing on every pixel, respectively. Figure \ref{fig:attention_maps_example} shows a visualization of attention-maps of 2 pixels.

\paragraph{Content-dependent attention-maps.}
The attention-map of the above mechanism is constructed by key $K$, query $Q$, and a relative positional encoding term. Both key and query are linear projections of input, while the relative positional encoding term also depends on the query \citep{bello2019attention}. Therefore, attention-maps are content-dependent. The multiplication between $K\in R^{HW \times d}$ and $Q^T \in R^{d \times HW}$ in Eq. \ref{eq:Att_wo} calculates the similarity between extracted features $K$ and $Q$. Its output $KQ^T$ with shape $HW \times HW$ indicates how similar each pixel's extracted feature is related to every other pixel. Finally, after adding a positional encoding and scaling, the final attention-maps are built.

However, as shown above, constructing attention-maps in such a way requires many learnable parameters and multiplication operations, and why the above mechanism is beneficial in the computer vision context is not obvious.

\section{Explicitly Modeled Attention-maps}

Intuitively, from the visualization of Figure \ref{fig:attention_maps_example}, it could be observed that the key $\times$ query highly depends on the input (Figure \ref{fig:normkq405}, \ref{fig:normkq592}), while the weights consisting of key, query and positional encoding (Figure \ref{fig:normkqp405}, \ref{fig:normkqp592}) does not depend much on the input content. The fact that the latter form is proved to increase the performance in image classification \citep{bello2019attention} indicates the importance of geometric information. This observation inspires us to design input independent attention-maps for vision tasks. Therefore, in comparison to previously described multi-head self-attention in the last section, we made the following modification: replace content-dependent attention-maps with content-independent explicitly modeled attention-maps using the assumption that neighbor pixels are more related than distant ones.

\subsection{General Form}

Generally, the spatial context for a given pixel within an image is mostly dominated by its neighbors, while more distant pixels have a minor contribution. Motivated by this observation and noting that attention-maps indicate the importance of all input pixels when focusing on each pixel, we explicitly design the weight distribution in attention-maps. In each attention-map of one specific pixel $i$, the weight assigned to any pixel $j$ decreases monotonically as the spatial distance between two pixels ($i$ and $j$) in input increases. Our proposed design for attention-maps is translational invariant and incorporates relative positional information %
by assigning spatial distance-dependent weights in a less costly manner than regular self-attention as introduced in previous section.

We define three steps of our attention mechanism as follows:

\begin{equation}
    \label{eq:V}
    V=X W_v
\end{equation}
\begin{equation}
    \label{eq:P}
    P = \Norm(G+1)V
\end{equation}
\begin{equation}
    \label{eq:ExpAtt}
    \ExpAtt(X)=PW^o
\end{equation}

In addition to existing notations, we define $d_v$ as the total number of channels of all attention heads, where $d_v=N \times d_v^h$.
$W_v\in R^{C \times d_v}$ is a learned linear transformation, which can be easily realized by $d_v$ $1D$ convolutions to create value V in $N$ heads together. We introduce the matrix $G\in R^{HW\times HW}$ in Eq. \ref{eq:P} to model weight distribution in attention-maps explicitly and constrain $G_{ij} \in (0,1]$. The concrete forms of $G$ are discussed in the subsequent section. We add an element-wise offset of 1 to $G$ to impose global feature consideration which ensures that information from the relatively distant area also has a reasonable magnitude of weight. This makes the whole spatial input considered. Following the normalization design of $Softmax$ in Eq. \ref{eq:Att_wo}, $\Norm$ denotes a row-wise normalization on input, and $\Norm(G+1)$ indicates the attention-maps. $\Norm$ normalizes values by directly dividing sum of values in each row of its input instead of additionally using an exponential function as in $Softmax$. This retains the effect of the designed $G$.

Additionally, we share attention-maps across heads. This helps to avoid splitting value V into multiple heads and concatenate them together in the end as Eq. \ref{eqn:multihead}. Finally, in Eq. \ref{eq:ExpAtt}, the output is linearly projected to achieve final self-attention output, where the weights are expressed in $W^o\in R^{d_v \times d_v}$.

\subsection{Fully Fixed Attention-maps \label{fully-fixed}}
A natural choice of modeling attention-maps explicitly is directly fixing all weights in attention-maps using exact functions. We design different simple functions for $G$ to model the weight decrease in attention-maps with increasing relative spatial distance. As a special case, we include the option of using a constant function. $G_{ij}$ indicates how much attention is paid to pixel j when focusing on pixel i. We denote $i_x, i_y, j_x, j_y$ as the corresponding spatial coordinate of pixel $i$ and $j$ in the input $X'$. For a fair comparison, all alternative formulations of $G$ are designed such that when $i=j$, $G_{ij}=G_{max}=1$ (one pixel should receive the highest attention when focusing on itself).

\textbf{Constant} In this case, the weights are distributed uniformly. Using different constants does not make a difference since the result after normalization will be the same.

\begin{equation}
    G_{ij}=1    
\end{equation}

\textbf{Linear decrease} For linear function, the weight decreases with an increasing Euclidean distance at a constant speed.

\begin{equation}
    G_{ij}=1-\frac{\sqrt{(i_x-j_x)^2+(i_y-j_y)^2}}{\sqrt{H^2+W^2}}    
\end{equation}

\textbf{Cosine decrease} In cosine decrease, weights are decreasing slower in the neighborhood and quicker in the middle distance and decreasing slower again in the very distant area.

\begin{equation}
    G_{ij}=0.5\cdot\left(1+\cos\left(\frac{\sqrt{\left(i_x-j_x\right)^2+\left(i_y-j_y\right)^2}}{\sqrt{H^2+W^2}}\pi\right)\right)
\end{equation}

The main advantage of fully fixed attention-maps is they can be fully pre-calculated before training and simply loaded into the model without introducing any learnable parameter or online computation. The implementation is very simple.

\subsection{Learnable Attention-maps}
Since our mechanism models weight distribution in attention-maps explicitly, how exactly the weights are distributed needs to be optimized. The optimized form of attention-maps can be tuned manually with different kinds of functions. Some examples include constant, linear, and cosine functions. However, manual tuning will cost lots of computing resources. Therefore, we further show some possible options to automatically determine the weight distribution in attention-maps.

\textbf{Gaussian}
First of all, $G$ is parametrized via the Gaussian kernel function defined as:

\begin{equation}
    G_{ij} = \exp\left( \frac{(\frac{i_x-j_x}{W})^2 + (\frac{i_y-j_y}{H})^2}{-2\sigma^2}\right)
\end{equation}

$\sigma$ is a learnable scalar, which indicates how flat the weights are distributed in attention-maps. Since the Gaussian kernel is also a radial basis function, we name $\sigma$ as radius, which is the shape parameter of this function. Visually, it also indicates the ''radius'' of the ''circle-shaped'' attention-maps, as shown in Figure \ref{fig:gaussiannorm405}.

\textbf{Exponential decrease}
We form the exponential decrease in two different ways, where the weight decreases with Euclidian and Manhatten distances, respectively as follows:

\begin{equation}
    G_{ij} = \exp\left( \frac{\sqrt{(\frac{i_x-j_x}{W})^2 + (\frac{i_y-j_y}{H})^2}}{-\sigma}\right)
\end{equation}
\begin{equation}
    G_{ij} = \exp\left( \frac{(\frac{|i_x-j_x|}{W}) + (\frac{|i_y-j_y|}{H})}{-\sigma}\right)
\end{equation}

Both options are similar to the Gaussian kernel function and mainly differ in the form of the numerator. In the choice of Gaussian kernel function, the weight also decreases exponentially. However, we keep them separate for convenience in later discussion.

\subsection{Analysis on Efficiency}

\begin{table*}
    \centering
\begin{small}
\begin{sc}
\begin{tabular}{ llll} 
 \hline
  & Parameters & Memory &Computation\\
  \hline
 Key and query & $N \times 2Cd $ & $O(N(HW)^2)$& $O(2NHWC^2d+Nd^2(HW)^2)$ \\
 Positional Encoding & $ 2(H+W-1)d$ & $O(HWd)$ &$O((HW)^2)$\\
 \hline
 Learnable att.-maps & 1 & $O((HW)^2)$ & $O((HW)^2)$\\
 Fully fixed att.-maps & 0&$O((HW)^2)$ & 0\\
\hline
\end{tabular}
\end{sc}
\end{small}
\caption{Comparison on parameter, memory and computation cost between explicitly modeled attention-maps and attention-maps calculated by key, query and relative positional encoding per layer.}
    \label{tab:memory}
\end{table*}
Since our method mainly improves the attention-maps construction, we focus the theoretical analysis on this part. Given input with height $H$, width $W$, and $C$ channels, the calculation of key and query in regular self-attention projects the input from $C$ channels to $d$ channels, respectively. Therefore they introduce $N \times 2Cd$ learnable parameters for $N$ heads in one layer. In each head, different attention-maps with shape $HW\times HW$ are saved, which has $O(N(HW)^2)$ memory cost. The projection of key and query costs $O(2HWC^2d)$ computation, and the multiplication between key and query costs $O(d^2(HW)^2)$ per head. The same computation is repeated in $N$ heads. Relative positional encoding introduces $ 2(H+W-1)d$ parameters with memory cost $O(HWd)$ and computation cost $O((HW)^2)$ per layer according to \citet{bello2019attention}.

However, using our explicitly modeled attention-maps will introduce only one learnable parameter (radius) if they are learnable and will not introduce any parameter if they are fixed. Further, we reduce the memory cost from $O(N(HW)^2)$ to $O((HW)^2)$ thanks to the sharing of attention-maps across heads. From the perspective of computation, fully fixed attention-maps can be completely pre-calculated and require no computation during training. Even our learnable variant only needs some scaling operations since the numerator of $G$ can be pre-calculated. Both variants of our method drastically increase the efficiency. The comparison is summarized in Table \ref{tab:memory}.

\section{Experiments}

In this section, we test our \textit{ExpAtt} module in widely used architectures such as ResNets \citep{he2016identity,he2016deep} and representative lightweight architecture such as MobileNetV2 \citep{Sandler2018MobileNetV2IR} on small scale and large scale image classification datasets including CIFAR10, CIFAR100 \citep{Krizhevsky2009LearningML}, Tiny ImageNet \citep{yao2015tiny} and ImageNet \citep{imagenet_cvpr09}. We report average accuracy for all experiments. The experiments show that our module leads to improvement in different architectures in multiple aspects. Since self-attention networks strongly motivate the proposed \textit{ExpAtt} module, we follow exactly the same network settings of \citet{bello2019attention} to integrate our \textit{ExpAtt} module into networks for comparability. Experiments in the same dataset use same data preprocessing.

\subsection{Integration by Feature Concatenation}

For an original convolution with stride 1 and output channels $C_{out}$ , we first split $C_{out}$ to standard convolution features $C_{conv}$ and \textit{ExpAtt} features $C_{expatt}$. In other words, $C_{conv}+C_{expatt}=C_{out}$. Subsequently, the convolution output has shape  $H \times W \times C_{conv}$ with $H$ and $W$ being the input height and width, respectively. The \textit{ExpAtt} output has shape $H \times W\times C_{expatt}$. From the perspective of multi-head self-attention, $C_{expatt}$ is equivalent to  $N \times d_v^h$, where $N$ is the number of attention heads, and $d_v^h$ is the depth of value in each head. 
Finally, the $ExpAtt$ output is concatenated with convolution output along channel dimension to receive the augmented convolutional features in shape $H\times W \times C_{out}$.  
For convolutions with stride 2, an additional $3\times3$ average pooling with stride 2 is applied to the \textit{ExpAtt} output to keep the spatial shape matching. 
The number of heads is fixed to 8 in ResNets and 4 in MobileNetV2. The ratio of $C_{expatt}/C_{out}$ is set to 0.1 for ResNets and 0.05 for MobileNetV2. When $C_{expatt}$ is not evenly dividable by 8 or 4, the closest value that is evenly dividable is taken.
Self-attention mechanism (including AA-Net and our ExpAtt-Net) is incorporated into $3\times 3$ convolutions of all 4 residual stages of ResNets in CIFAR and only last 3 stages of ResNets in ImageNet experiments. The integration into MobileNetV2 starts when channel number is $24 \times 6$ through concatenation with $1\times 1$ convolutions. More details are offered in the Appendix.

\subsection{Training}
Models are trained from scratch. All experiments (including AA-Net and ExpAtt-Net) are based on the respective baselines from PyTorch \citep{Paszke2019PyTorchAI}, use synchronous SGD with momentum 0.9, and cosine learning rate with restarts \citep{loshchilov2016sgdr} for in total 450 epochs, 164 epochs, and 324 epochs in CIFAR, ImageNet, and Tiny ImageNet experiments respectively. Concretely, in the first 15 epochs, learning rate is linearly increased to 0.05, than a cosine learning rate with restarts at 25,45,85,165,325 epochs is applied where appliable. Additionally, CIFAR experiments use learning rate 0.0002 between epoch 325 and 450. Batch size of all experiments are chosen to fit the GPU memory. The radius $\sigma$ of Gaussian kernel is initialized to 0.75. 

\subsection{ResNet50 in CIFAR-10,100}

Tab. \ref{tab:cifar100} shows the performance of Resnet50 when the attention-maps of our \textit{ExpAtt} module are parametrized differently with various simple functions. To fit the resolution of CIFAR, we remove the first average pooling and change the stride of the first convolution to one in ResNet50. All considered attention modules with different parametric attention-maps outperform the ResNet50 baseline and the plain self-attention in AA-ResNet50. All functions perform similarly with the Gaussian-kernel being slightly better than others. This may occur because different functions have similar gaussian-like patterns. Surprisingly, even using uniform distribution in attention-maps achieves competitive performance compared to AA-ResNet50. This may mean regularization is helpful in the attention module.

\begin{table}
    \centering
\begin{small}
\begin{sc}

\begin{tabular}{ p{1.9cm}*{3}{p{.75cm}} } 
 \hline
 Decay func.&Para. & FLOPs & Acc. \\
 \hline
 \multicolumn{4}{c}{Cifar-10}\\
 ResNet50 & 23.7M &1.31G& 90.20 \\ 
 AA-ResNet50  & 23.9M &1.45G& 90.78\\
 Uniform & 22.7M& 1.25G&90.77 \\
 Cosine  & 22.7M & 1.25G&90.96 \\
 Linear  & 22.7M&1.25G&90.91 \\
 Exp. Euclid. & 22.7M & 1.25G&90.94 \\
 Exp. Man. & 22.7M & 1.25G&\textbf{91.02} \\
 Gaussian & 22.7M &1.25G& \textbf{90.99} \\ 
 \hline
 \multicolumn{4}{c}{Cifar-100}\\
 ResNet50  & 23.7M & 1.31G& 79.46 \\
 AA-ResNet50 & 23.9M &1.45G & 80.32 \\
 Cosine & 22.7M &1.25G& 80.74 \\ 
 Uniform & 22.7M &1.25G& 80.76 \\
 Linear & 22.7M &1.25G& 80.82 \\
 Exp. Euclid. & 22.7M &1.25G& 80.90 \\
 Exp. Man. & 22.7M &1.25G& 80.91 \\
 Gaussian & 22.7M &1.25G & \textbf{81.02}\\
 \hline
\end{tabular}
\end{sc}
\end{small}
\caption{
	Performance of modified ResNet50 using different parametric attention-maps on CIFAR-10/100.
}
    \label{tab:cifar100}
\end{table}
\begin{table}
\centering
\begin{small}
\begin{sc}
\begin{tabular}{l*{4}{p{.75cm}} } 
 \hline
 Type & Para. & FLOPs& Top1 &Top5\\
 \hline
 ResNet34 & 21.8M & 3.6G&73.30 & 91.42 \\
 SE-ResNet34 & 22.0M & 3.6G&74.30 & 91.80 \\
 CBAM-ResNet34& 22.0M &3.7G &74.01 &91.76 \\
 AA-ResNet34 & 20.7M & 3.6G&\textbf{74.70} & \textbf{92.00}\\
\emph{Gaussian-ExpAtt} & 17.3M & 3.1G&74.24  & 91.81\\ 
 \hline
ResNet50 & 25.6M & 3.8G &76.15 & 92.87 \\
SE-ResNet50 & 28.1M & 3.9G &77.50& 93.70 \\
CBAM-ResNet50 & 28.1M&3.9G &77.34  & 93.69\\
ECA-ResNet50 & 24.4M & 3.9G & 77.48 &93.68\\
AA-ResNet50 & 25.8M & 4.2G &77.70 & 93.80\\
\emph{Gaussian-ExpAtt}& 24.5M & 4.0G& \textbf{78.13} & \textbf{94.07}\\ 
\hline
ResNet101 &44.5M & 7.6G&77.37 &93.56\\
SE-ResNet101 & 49.3M & 7.6G&78.40 & 94.20 \\
CBAM-ResNet101 & 49.3M&7.6G &78.49 &94.31 \\
ECA-ResNet101 &42.5M &7.4G & 78.65 &94.34\\ 
AA-ResNet101 & 45.4M & 8.1G&78.70 & 94.40\\
\emph{Gaussian-ExpAtt} & 42.7M & 7.6G& \textbf{79.56} & \textbf{94.78}\\ 
\hline
ResNet152 & 60.2M & 11.3G&78.31 & 94.06\\
SE-ResNet152 & 66.8M & 11.3G& 78.90 & 94.50 \\
ECA-ResNet152 & 57.4M&10.8G &78.92&94.55\\
AA-ResNet152 & 61.6M & 11.9G&79.10  & 94.60\\ 
\emph{Gaussian-ExpAtt} & 57.6M & 11.1G& \textbf{80.02} & \textbf{94.85}\\ 
 \hline
\end{tabular}
\end{sc}
\end{small}
\caption{
	Performance of ResNets utilizing different attention modules in ImageNet. Our methods are cursive.
}
    \label{tab:ImageNet}
\end{table}

\subsection{ResNets in ImageNet}
In table \ref{tab:ImageNet}, we compare representative networks exploring channel attention (SE-Net\citep{hu2018squeeze}),
efficient channel attention (ECA-Net\citep{Wang2019ECANetEC}), channel and spatial attention independently (CBAM \citep{woo2018cbam}) and jointly (AA-Net \citep{bello2019attention}) with our \textit{ExpAtt-Net}, which explores efficient joint channel and spatial attention.
The methods are compared by integration into multiple Resnet architectures.
The Gaussian kernel parametrized \textit{ExpAtt} improves its counterpart using key and query (AA-ResNet) by 0.47\%, 0.87\%, and 0.92\% on ResNet50, ResNet101, and ResNet152, respectively in Top1 accuracy, which indicates that increasing architecture depth is beneficial for explicit modeling of the attention-maps. This may due to the regulation introduced by \textit{ExpAtt} model, which helps to decrease training difficulty when the model becomes deeper. Compared to ECA-ResNet, which is designed to execute efficient channel attention, we use a similar number of parameters and GFLOPs while achieving a higher Top1 accuracy. This further proves the benefit of our method. 

However, the \textit{ExpAtt} augmented ResNet34 underperforms the AA-Resnet34, though still outperforming ResNet34 baseline. Since ResNet50/101/152 uses a different type of residual block (bottleneck) than ResNet34, this may indicate that our method is more suitable to model attention weight distribution in architectures with bottleneck residual block which consists of a $1\times1$, $3\times3$, and $1\times1$ convolution instead of the residual blocks consisting of two $3\times3$ convolutions (e.g., ResNet34).

\subsection{MobileNetV2 in Tiny ImageNet}

In this section, we use a parameter efficient architecture MobileNetV2 as the backbone and compare \textit{ExpAtt} with AA-Net \citep{bello2019attention} using precisely the same training and network setting. Following \citet{bello2019attention}, we apply the Gaussian parametrized \textit{ExpAtt} module in an inverted bottleneck by replacing part of expansion point-wise convolution channels. Table \ref{MobileNetV2} shows that we achieve an accuracy improvement with lower model complexity compared to AA-MobileNetV2. Since the inverted bottleneck mainly consists of two point-wise and one depth-wise convolution, this also suggests a way to let our method complement depth-wise convolution.

\begin{table}[t]
    \centering
\begin{sc}
\begin{tabular}{ lccc } 
\hline
Architecture & Para.& FLOPs & Acc.\\
\hline
MobileNetV2 & 3.50M& 0.32G& 64.72\\ 
AA-MobileNetV2&3.55M&0.33G&65.89\\
ExpAtt-MobileNetV2 &3.51M &0.32G&\textbf{66.14}\\
\hline
\end{tabular}
\end{sc}
\caption{Performance of MobileNetV2 and its variants augmented with self-attention modules.}
\label{MobileNetV2}

\end{table}

\subsection{Ablation Study}

\paragraph{Sharing attention-maps across heads} From the perspective of multi-head self-attention, each head can have a different set of attention-maps. Therefore, we study the effect of sharing and not sharing attention-maps (parametrized by Gaussian kernel) across heads by experiments in ResNet50 on CIFAR100. The result shows that sharing attention-maps achieves 81.02\% Top1 accuracy while using different attention-maps across heads achieves only 80.78\%. One possible explanation is that the tied radius parameter might reduce the difficulty of joint optimization in training neural networks.

\paragraph{Interplay with content-based method} In this study, we try to understand whether our content-independent method is orthogonal to the content-dependent one using ResNet50 on CIFAR100. In all augmented layers, we combine the two methods by
concatenating the output of \textit{ExpAtt} and the content-based method's output \citep{bello2019attention}. Unfortunately, the accuracy (78.19\%) is worse than plain ResNet50 (79.46\%). The result suggests that both methods are not complementary, though they individually have an obvious improvement over vanilla networks. 

\paragraph{Importance of global features} Though our method is motivated by focusing local features, we explicitly constrain the attention-maps to consider global features by using element-wise plus one to $G$. Without using element-wise plus one in Eq. \ref{eq:P}, attention-maps parametrized by Gaussian kernel achieves only $80.42\%$ in CIFAR100 while additionally using element-wise plus one achieves $81.02\%$. This indicates the importance of considering global features.

\subsection{Results of Learned Radius}

\begin{figure}[t]
\centering
\begin{subfigure}{.5\textwidth}
    \centering
    \includegraphics[width=.8\linewidth]{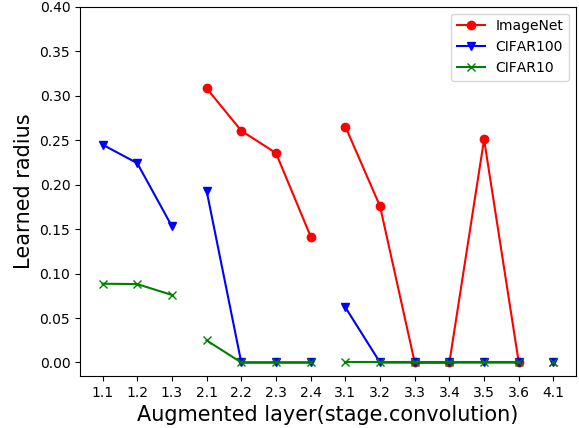}
    \caption{}%
    \label{fig:learned_sigma50}
\end{subfigure}
\begin{subfigure}{.5\textwidth}
    \centering
    \includegraphics[width=.8\linewidth]{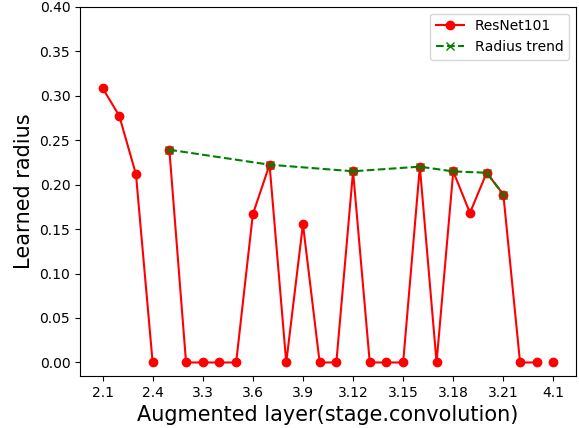}
    \caption{}%
    \label{fig:learned_sigma101}
\end{subfigure}
\caption{The final results of learned radius in different augmented layers of Resnet50 on CIFAR and ImageNet (a) and ResNet101 on ImageNet (b). Results in different stages are split for clarity.
}
\label{fig:learned_sigma}

\end{figure}

Since radius is the only learnable parameter of Gaussian kernel parametrized \textit{ExpAtt}, we show how it varies between layers. The augmented layers are denoted as ''stage.convolution'' on the x-axis, because ResNet50 consists of several stages, and each stage has several $3\times 3$ convolution layers.
Fig. \ref{fig:learned_sigma50} shows the final learned value of the radius in different layers of ResNet50 on different datasets. 
There is an apparent decreasing trend in the learned radius as the layer depth increases in the first augmented stage. This suggests that the global context is more important in early layers than later ones. Further, the radius learned in stage one on CIFAR100 is higher than the learned radius on CIFAR10. This may mean that the global context is more important as the prediction task becomes more difficult.
In the third stage, although the radius is not continuously decreasing, the general trend of ''peak radius'' in neighbors is still decreasing as the layer goes deeper. This trend can be easier observed in deeper network such as ResNet101 (Fig. \ref{fig:learned_sigma101}) .
In the fourth stage, only the first augmented convolution is parametrized by the Gaussian kernel, and the final radius is close to zero in all datasets. Radius close to zero means that attention-map of any pixel F assigns the highest weight to pixel F and considers information from all other surrounding pixels equally.

\section{Conclusion}

We aim at offering an efficient self-attention mechanism for vision task. To this end, we propose an \textit{ExpAtt} module to explicitly model weight distribution in attention-maps by incorporating a geometric prior. Despite the simplicity of this module compared to self-attention, experimental results show that it improves the performance of multiple architectures, including widely used ResNets and lightweight MobileNetV2. Surprisingly, it outperforms the regular self-attention design not only in efficiency but also in accuracy when integrated into the bottleneck residual block. Although experiments focus on image classification, we expect \textit{ExpAtt} to be applicable to other vision tasks, because the assumption that nearby pixels are more related than distant ones is a general principle in images and that the Resnet baseline for image classification is widely used as backbone for many other tasks.

\section*{Acknowledgements}
The authors would like to thank Zhongyu Lou from Robert Bosch GmbH for insightful comments and discussions.

\section*{Ethical Impact}

Our content-independent attention-maps can help to decrease bias introduced by datasets against minorities. For example, in a face image recognition task, more training images of majorities may cause the system to perform worse in people of minorities. Self-attention with content-dependent attention-maps might learn such bias due to its large number of parameters learned from datasets. In contrast, both variants of our method try to avoid learning such bias. Concretely, the fully fixed attention-maps will not be influenced by any bias of datasets because it learns nothing from datasets. Our learnable attention-maps would also be much less influenced by such bias compared to content-based methods, since it only learns a single shape parameter thanks to our very general assumption that neighbor pixels are more related than distant ones. However, we still note that the system might be misused to cause negative ethical impact.

\bibliography{aaai2021}
\bibliographystyle{aaai21}

\end{document}